\documentclass[conference]{IEEEtran}
\IEEEoverridecommandlockouts
\usepackage{cite}
\usepackage{amsmath,amssymb,amsfonts}
\usepackage{transparent}
\usepackage{algpseudocode}
\usepackage{graphicx}
\usepackage{textcomp}
\usepackage{xcolor}
\usepackage{soul}
\usepackage{pgfplots}
\usepackage{hyperref}
\def\BibTeX{{\rm B\kern-.05em{\sc i\kern-.025em b}\kern-.08em
    T\kern-.1667em\lower.7ex\hbox{E}\kern-.125emX}}

\begin{document}

\title{An Approach for Air Drawing Using Background Subtraction and Contour Extraction}

\author{\IEEEauthorblockN{Ramkrishna Acharya}
\IEEEauthorblockA{\textit{Master in Data Science} \\
\textit{Friedrich-Alexander-Universität Erlangen-Nürnberg}\\
Erlangen, Germany \\
qramkrishna.acharya@fau.de}
}

\maketitle

\begin{abstract}
In this paper, we propose a novel approach for air drawing that uses image processing techniques to draw on the screen by moving fingers in the air. This approach benefits a wide range of applications such as sign language, in-air drawing, and 'writing' in the air as a new way of input. The approach starts with preparing ROI (Region of Interest) background images by taking a running average in initial camera frames and later subtracting it from the live camera frames to get a binary mask image. We calculate the pointer's position as the top of the contour on the binary image. When drawing a circle on the canvas in that position, it simulates the drawing. Furthermore, we combine the pre-trained Tesseract model for OCR purposes. To address the false contours, we perform hand detection based on the haar cascade before performing the background subtraction. In an experimental setup, we achieved a latency of only 100ms in air drawing. The code used to this research are available in GitHub as \href{https://github.com/q-viper/Contour-Based-Writing}{https://github.com/q-viper/Contour-Based-Writing}.

\end{abstract}


\section{Introduction}
Air Drawing - an approach to drawing on screen by moving the hand in the air has the potential to aid in gesture recognition and gesture-based writing applications. One such remarkable model is proposed by Y. Yin et al.\cite{AirContour} where authors used a smartwatch to collect sensor data. Using air drawing based on background subtraction and contours can be the key technology to facilitate gesture-based interaction in the air and can be used in many scenarios. For example, this could be a fun yet good way for children to learn how to write and draw. Furthermore, it could be a possible replacement for input devices for on-screen text, digital drawing, or painting. The main issue in Air Writing is finding the right place to draw. It can be calculated using gesture recognition and hand detection. And the use of sensors can be challenging in terms of the cost of the device, configuring the device, portability, and ease of use of sensor data and processing the data.

In order to address the aforementioned issues with sensor-based devices and approaches, we propose a novel image processing-based method. We use a stable web cam for 2D image frames, and prepare background images on an ROI. We perform threshold on the image generated from background subtraction \cite{backgroundSubtractionStudy} from the live ROI image to get the binary image. It is further used to compute the contours. The top of the contour with the largest area is called a pointer in this paper. The use of haar cascade-based \cite{haarCascade} hand detection model is used to check if the hand is detected in ROI or not.

Furthermore, we integrate our model with Tesseract \cite{tesseract} for OCR operation. In this part, we pass our drawing to a pre-trained Tesseract and get a present. 

\section{Methods}
\subsection{Architecture and System Design}
We acquire two simultaneously running images concatenated horizontally to form one. In the first image, we show the live camera frames as well as two ROI boxes on the bottom right. In the early frames of the running, we calculate the running average background image for the individual ROI. We use the first ROI box for drawing mode selection and the second for moving pointers.
In the second image, we show the movement of the pointer, drawing, and a novel term VUI (Visual User Interface) where we show different UI modes and operations to aid the drawing procedure as described in Table ~\ref{tab:table}.

\begin{center}
\begin{figure}[htbp]
\centering
\label{HasaRelation}
\def\svgwidth{0.5\textwidth}
%% Creator: Inkscape 1.2.2 (732a01da63, 2022-12-09), www.inkscape.org
%% Accompanies image file 'drawing1.pdf' (pdf, eps, ps)
\begingroup%
  \makeatletter%
  \providecommand\color[2][]{%
    \errmessage{(Inkscape) Color is used for the text in Inkscape, but the package 'color.sty' is not loaded}%
    \renewcommand\color[2][]{}%
  }%
  \providecommand\transparent[1]{%
    \errmessage{(Inkscape) Transparency is used (non-zero) for the text in Inkscape, but the package 'transparent.sty' is not loaded}%
    \renewcommand\transparent[1]{}%
  }%
  \providecommand\rotatebox[2]{#2}%
  \newcommand*\fsize{\dimexpr\f@size pt\relax}%
  \newcommand*\lineheight[1]{\fontsize{\fsize}{#1\fsize}\selectfont}%
  \ifx\svgwidth\undefined%
    \setlength{\unitlength}{591.749282bp}%
    \ifx\svgscale\undefined%
      \relax%
    \else%
      \setlength{\unitlength}{\unitlength * \real{\svgscale}}%
    \fi%
  \else%
    \setlength{\unitlength}{\svgwidth}%
  \fi%
  \global\let\svgwidth\undefined%
  \global\let\svgscale\undefined%
  \makeatother%
  \begin{picture}(1,0.51249507)%
    \lineheight{1}%
    \setlength\tabcolsep{0pt}%
    \put(0,0){\includegraphics[width=\unitlength,page=1]{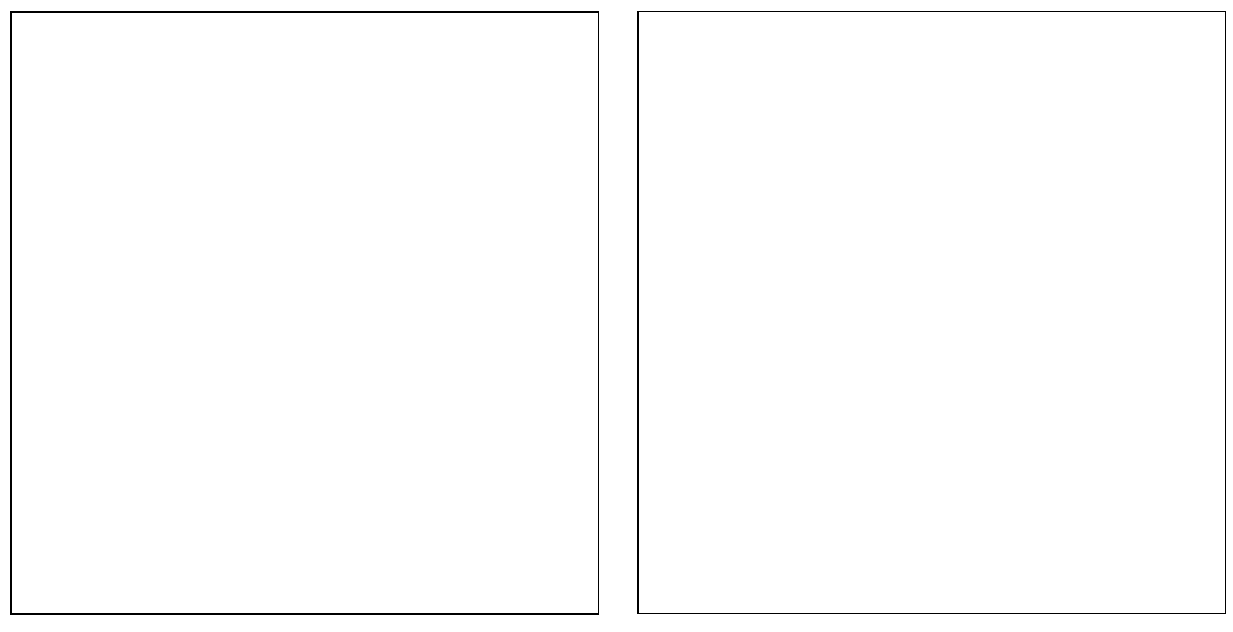}}%
    \put(0.19239702,0.14053377){\color[rgb]{0,0,0}\makebox(0,0)[lt]{\lineheight{1.25}\smash{\begin{tabular}[t]{l}Drawing ROI\end{tabular}}}}%
    \put(0,0){\includegraphics[width=\unitlength,page=2]{drawing1.pdf}}%
    \put(0.52347992,0.24834199){\color[rgb]{0,0,0}\makebox(0,0)[lt]{\lineheight{1.25}\smash{\begin{tabular}[t]{l}Drawing Window\end{tabular}}}}%
    \put(0.58651909,0.46152341){\color[rgb]{0,0,0}\makebox(0,0)[lt]{\lineheight{1.25}\smash{\begin{tabular}[t]{l}Modes VUI\end{tabular}}}}%
    \put(0.19845346,0.26510097){\color[rgb]{0,0,0}\makebox(0,0)[lt]{\lineheight{1.25}\smash{\begin{tabular}[t]{l}Modes ROI\end{tabular}}}}%
    \put(0.05302867,0.45541182){\color[rgb]{0,0,0}\transparent{0}\makebox(0,0)[lt]{\lineheight{1.25}\smash{\begin{tabular}[t]{l}Live Camera Feed  \end{tabular}}}}%
    \put(0,0){\includegraphics[width=\unitlength,page=3]{drawing1.pdf}}%
  \end{picture}%
\endgroup%

\caption{Left: Image from Webcam where ROI drawing happens. Right: A canvas image where drawing, VUI, and pointer movement happen. The dimension of both is (420, 720) pixels. }
\end{figure}
\end{center}

% \begin{center}
%  \centering
%  \includegraphics[width=0.5\textwidth]{drawing1.png}
%  \label{fig:guiconcept}
%  \captionof{Fig. 1. }{Left image is from Webcam where ROI drawing happens and the right image is a canvas where drawing, VUI, and pointer movement happen. }
% \end{center}

% \section{Methods}
\subsection{Preparation of Background Image for Individual ROIs}
Prepared background image plays a crucial part in contour extraction as it is used in background subtraction. Hence, we prepare it by taking a running average of ROIs for as many frames as possible. We are calling the average image the background image in this research. In our experiment, we found that taking an average of the first 100-150 frames gives a binary image with the least noise.
\begin{equation}\label{eq:running_avg}
dst(x, y) = (1-\alpha) * dst(x,y) + \alpha *  src(x, y)
\end{equation}

Where,
\begin{itemize}
  \item \textit{dst} is the destination image or accumulator
  \item \textit{src} is the source image
  \item \begin{math}\alpha \end{math} is the weight of the input image, default 0.8
\end{itemize}

\subsection{Pointer Position Calculation From ROI Binary Images}
We pass ROI images through a hand detector before doing background subtraction and if detected, calculate the potential position of the pointer.

By subtracting the aforementioned ROI background image from the ROI image, we could get a residual image on which we apply the binary threshold. We find the contours in the residual image. Contours in image processing are the curves or boundaries formed by interconnected pixels of similar color. The position of the pointer was taken as the top position of the contour which has the highest area. Since this position is calculated on the ROI scale, we transformed it into the scale of a drawing image.

Let $(r_{x0}, r_{y0})$ be top left and $(r_{x1}, r_{y1})$ be bottom right coordinates of ROI, $(f_{x0}, f_{y0})$ be top left and $(f_{x1}, f_{y1})$ be bottom right coordinates of the frame, and $(r_x, r_y)$ be pointer position in ROI, then the pointer position in canvas $(c_x,c_y)$ are calculated as follows:
\begin{equation}
    P_c = (c_x, c_y) = (\frac{f_{x1}-f_{x0}}{r_{x1}-r_{x0}} * r_x, \frac{f_{y1}-f_{y0}}{r_{y1}-r_{y0}} * r_y)
\end{equation}

\subsection{Pointer and its Modes}
To simulate the pointer movement, we draw a circle on a temporary image on position $P_c$ on every frame then is added to the canvas image. On the position of the $P_c$ in the canvas image, we perform operations based on the selected mode. We choose white as a background color of the canvas where some user interfaces are also shown and upon hovering them, operation mode activates.
Table 1 contains all the modes we experimented with.
\begin{table}[hbt!]
 \caption{Pointer Operation Modes}
  \centering
  \begin{tabular}{ll}
    % \toprule
    % \multicolumn{2}{c}{Part}                   \\
    % \cmidrule(r){1-2}
    \hline
    Name     & Description      \\
    \hline
    % \midrule
    Clear & Clear the canvas to the default state.  \\
    Color   & Change the pointer's drawing color. \\
    Detect  & Perform OCR using Tesseract \cite{tesseract}      \\
    Draw    & Change the draw color to the currently selected color.\\
    Erase   & Change the draw color to the background color.\\
    % Exit    & Exit the entire system.\\
    Move    & Change cursor mode to move only.\\
    % Restart & Restart the system. \\
    Save    & Saving the draw/writing.\\
    
    % \midrule
    % \bottomrule
    \hline
  \end{tabular}
  \label{tab:table}
\end{table}

\subsection{Experiments and Running System}
\begin{figure}
 \centering
 \includegraphics[width=0.5\textwidth]{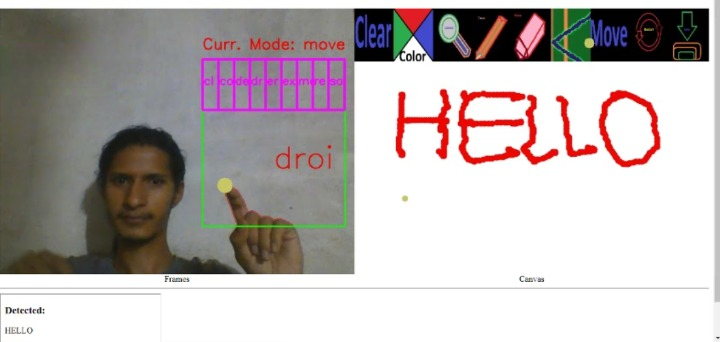}
 \caption{Running Air drawing, mode selection, and detection.}
\label{fig:airdraw}
 \end{figure}

Our experiments were done using the Dell 3543, I5 5th generation processor with 8GB RAM, a webcam of 28FPS, and a frame size of (420, 720), and Figure \ref{fig:airdraw} shows the working state of our experiment. 

\section{Results and Discussion}
Unlike sensor-based devices, which calibrate 3D gesture to 2D contour, our approach uses image processing for contour extraction which is a comparatively simple, computationally cheap, and minimal-effort process. Our method has an average latency of 100ms which beats the sensor-based method proposed by Y. Yin et al. \cite{AirContour} by 50ms. However, this duration is positively correlated to the frame dimension and the noise in the environment. Furthermore, we found that 98\% of the detected texts were correct. Apparently, detection is related to the size of the pointer and it also adds some delay in inter-result duration. In addition to that, found 4 pixels of pointer diameter to be best for (420, 720) frame dimensions.

Despite our approach having a simple simpler, faster, and cheaper way of contour extraction and air drawing, it comes short with a noisy environment and moving camera. However, the solution to hand tracking on moving cameras using hand landmark detection is publicly available by Mediapipe \cite{mediapipe1}.

\section{Conclusion}
In this study, we used novel image processing techniques to perform air drawing and found that it has the potential to become a simpler, cheaper, and faster alternative to sensor-based methods. We conclude that by combining state-of-the-art models like image completion, our approach has the potential to open new applications for air drawing. 

\bibliographystyle{IEEEtran}
\bibliography{references}

\begin{thebibliography}{1}
\providecommand{\url}[1]{#1}
\csname url@samestyle\endcsname
\providecommand{\newblock}{\relax}
\providecommand{\bibinfo}[2]{#2}
\providecommand{\BIBentrySTDinterwordspacing}{\spaceskip=0pt\relax}
\providecommand{\BIBentryALTinterwordstretchfactor}{4}
\providecommand{\BIBentryALTinterwordspacing}{\spaceskip=\fontdimen2\font plus
\BIBentryALTinterwordstretchfactor\fontdimen3\font minus
  \fontdimen4\font\relax}
\providecommand{\BIBforeignlanguage}[2]{{%
\expandafter\ifx\csname l@#1\endcsname\relax
\typeout{** WARNING: IEEEtran.bst: No hyphenation pattern has been}%
\typeout{** loaded for the language `#1'. Using the pattern for}%
\typeout{** the default language instead.}%
\else
\language=\csname l@#1\endcsname
\fi
#2}}
\providecommand{\BIBdecl}{\relax}
\BIBdecl

\bibitem{AirContour}
Y.~Yin, L.~Xie, T.~Gu, Y.~Lu, and S.~Lu, ``Aircontour: Building contour-based
  model for in-air writing gesture recognition,'' \emph{ACM Transactions on
  Sensor Networks}, vol.~1, no.~1, January 2019.

\bibitem{backgroundSubtractionStudy}
Y.~Benezeth, P.~Jodoin, B.~Emile, H.~Laurent, and C.~Rosenberger, ``Comparative
  study of background subtraction algorithms,'' \emph{Journal of Electronic
  Imaging}, vol.~19, 2010.

\bibitem{haarCascade}
P.~Viola and M.~Jones, ``Rapid object detection using a boosted cascade of
  simple features,'' in \emph{Proceedings of the 2001 IEEE Computer Society
  Conference on Computer Vision and Pattern Recognition (CVPR 2001)}, Kauai,
  HI, USA, 2001.

\bibitem{tesseract}
C.~Patel, A.~Patel, and D.~Patel, ``Optical character recognition by open
  source ocr tool tesseract: A case study,'' \emph{International Journal of
  Computer Applications}, vol.~55, pp. 50--56, October 2012.

\bibitem{mediapipe1}
\BIBentryALTinterwordspacing
Mediapipe, ``Mediapipe hands,'' Online, 2023, accessed: 2023-02-01. [Online].
  Available: \url{https://google.github.io/mediapipe/solutions/hands.html}
\BIBentrySTDinterwordspacing

\end{thebibliography}
\vspace{12pt}
\end{document}